# Weakly Supervised Object Detection with Pointwise Mutual Information

Rene Grzeszick, Sebastian Sudholt, Gernot A. Fink
TU Dortmund University, Germany
{rene.grzeszick,sebastian.sudholt,gernot.fink}@tu-dortmund.de

## Abstract

*In this work a novel approach for weakly supervised object detection that incorporates pointwise mutual information is presented. A fully convolutional neural network architecture is applied in which the network learns one filter per object class. The resulting feature map indicates the location of objects in an image, yielding an intuitive representation of a class activation map. While traditionally such networks are learned by a softmax or binary logistic regression (sigmoid cross-entropy loss), a learning approach based on a cosine loss is introduced. A pointwise mutual information layer is incorporated in the network in order to project predictions and ground truth presence labels in a non-categorical embedding space. Thus, the cosine loss can be employed in this non-categorical representation. Besides integrating image level annotations, it is shown how to integrate point-wise annotations using a Spatial Pyramid Pooling layer. The approach is evaluated on the VOC2012 dataset for classification, point localization and weakly supervised bounding box localization. It is shown that the combination of pointwise mutual information and a cosine loss eases the learning process and thus improves the accuracy. The integration of coarse point-wise localizations further improves the results at minimal annotation costs.*

## 1. Introduction

The classification and localization of objects is one of the main tasks for the understanding of images. Much progress has been made in this field based on the recent developments in Convolutional Neural Networks (CNNs) [12]. The error rates in prominent tasks like ImageNet competitions have been reduced by a large margin over the last five years [16]. While the models become more and more powerful, the required data can still pose a bottleneck. In many localization tasks very detailed annotations are required in order to train a visual detector. Typically, an annotation is required that has the same level of detail as the desired output of the detector, e.g. bounding boxes or even pixel level annotations.

As obtaining these annotations is expensive, weakly supervised learning approaches become of broader interest. These methods require a lower level of supervision during training. An object detector can be trained while only labeling images with respect to the presence or absence of certain objects. Similar to supervised tasks, great progress has been made in the field of weakly supervised learning by incorporating CNNs.

State-of-the-art approaches in weakly supervised object detection use region proposals in order to localize objects. They evaluate these proposals based on a CNN that is solely trained on image level annotations. In [2] images are evaluated based on a pre-trained classification network, i.e. a VGG16 network that is pre-trained on ImageNet. The convolutional part of this network is evaluated on a given image, computing a feature map. Based on heuristic region proposals, e.g. from selective search or edge boxes, a set of candidate regions is cropped from the feature maps. These candidates are then processed by a Spatial Pyramid Pooling (SPP) [9] layer which is followed by fully connected layers. Only this part is trained in a weakly supervised fashion based on image level annotations and the relation between different candidate regions. In [10] a similar approach is followed. Here, two different loss functions are introduced which incorporate either an additive or subtractive center surround criterion for each candidate region. It has been shown that this allows for improving the results compared to [2]. While these approaches show state-of-the-art performance, the incorporation of region proposals often comes at a high computational cost and is based on heuristic design decisions and expert knowledge (cf. [7]).

It has also been shown that with a deeper understanding of CNNs and their activations, the visualization of important filters can be leveraged for localizing objects. These approaches do not include additional region proposals so that they can be learned in an end-to-end fashion. A comparison of recent visualization approaches can be found in [17]. In [14] and [19] CNNs are trained on image level annotations for the task of object detection. The work in [14] applies max pooling for predicting the locations of objects in a weakly supervised manner. A multi-scale training approach is employed in order to learn the locations more accurately. The network training is performed using a binary logistic loss function (also known as cross-entropy loss) which in turn allows to predict a binary vector indicating the presence of multiple objects at once. A similar approach is followed in [19], but in contrast to [14], a global average pooling followed by a softmax is applied. In [19], it is argued that the global average pooling captures the extent of an object rather than just a certain part of an object. Based on the global average of the last filter responses, weights are computed which calculate the importance of each filter for a certain class. This can then be used in order to highlight the presence of certain classes in a so-called class activation map (CAM). These class specific activations show an object's extent and can therefore be leveraged in order to predict objects in a weakly supervised manner.

Besides different approaches for weakly supervised learning, there are a few methods that deal with learning from annotations which require a minimal annotation effort. In [11] CAMs are improved by adding micro annotations. Similar regions are grouped and manually labeled in order to remove false positive detections and obtain a more accurate localization. For example, trains are consistently co-occurring with tracks and thus often falsely recognized in the localization. In [1], point-wise annotations are introduced for the task of semantic segmentation. These provide a coarse localization of objects that also comes with a low annotation effort. It has been shown that the additional effort for point wise annotations is as low as approx. 2.1 sec. per image compared to image level annotations [1]. Such annotations may also provide an interesting cue of information to boost the performance of weakly supervised object detectors.

Another interesting aspect of weakly supervised learning with CNNs are the loss functions. For example, in [14] a binary logistic loss function is used. Most prominently this loss is also used in tasks where multiple entities are predicted at once, as, for example, in attribute prediction [3, 8]. In [8], multiple scene attributes are recognized simultaneously in a CNN. It is shown that this approach outperforms traditional per attribute recognizers which are typically SVMs on top of heuristic feature representations or later on SVMs trained on CNN features [15, 20]. The simultaneous prediction is important as training multiple deep networks for each attribute is not suitable. Furthermore, the larger number of samples is an advantage for training. It can be assumed that the network also learns which attributes are typically appearing simultaneously within its feature representations. This idea is followed in [3], where an embedding is computed which encodes the mutual information between two attributes in the data, the pointwise mutual information (PMI) embedding. A CNN is trained based on the feature vectors of the embedding space. The predictions are then also made with respect to the embedding space. Given that this is a continuous, non-categorical, space, traditional softmax or binary logistic loss functions can no longer be used. Thus, a cosine loss function is employed for training the CNN. However, since the predictions are made in the embedding space, the presence of certain attributes can no longer be predicted in a straightforward manner. They are thus predicted using the cosine similarity between the networks output and vectors with a one-hot encoding that indicate the presence of a certain attribute.

In this work a fully convolutional network that incorporates PMI for weakly supervised object detection is introduced. The network learns exactly one filter per object class and does not incorporate additional information such as region proposals. The contributions are as follows: The network incorporates a PMI embedding layer and is trained through a cosine loss, yet is still able to predict the presence of objects in a single forward pass. It is furthermore shown how to integrate annotations for either image level annotations or point-wise annotations using a SPP layer.

## 2. Method

A fully convolutional network architecture is proposed which allows for object detection. An overview is given in Fig. 1. The network is designed to learn exactly one filter for each object class (see sec. 2.1). These filters are then followed by a SPP layer which allows for training the network in a weakly supervised fashion (cf. [14, 19]). For example, image level annotations correspond to the first level of the SPP, whereas coarse localizations can be encoded by using multiple levels which indicate the presence of an object in a certain region (as described in sec. 2.2). In order to account for co-occurrences of objects, an integrated learning step is proposed. A PMI layer, more precisely the positive pointwise mutual information (PPMI), is included in the network which projects the object prediction scores into a de-correlated feature space (see sec. 2.3). As the features to be learned in this feature space are in $\mathbb{R}^n$ and non-categorical, the cosine loss function is applied for training. The error is backpropagated through the network, including a backprojection of the PMI transformation so that the network still outputs scores for the presence of objects at the last convolutional feature map.

### 2.1. Fully Convolutional Network

The proposed fully convolutional network architecture is similar to many other fully convolutional networks and based on the VGG networks [18]. Here, the fully connected layers of the VGG16 architecture are replaced by two additional convolution layers: one with 512 filters and one with exactly one filter per object class. Thus, instead of learning a global mapping of filters to object classes as in the CAM approach (cf. [14, 19]), the network learns exactly one filter which is responsible for indicating the presence of an object class at a certain location. This behavior is rather similar to networks for semantic segmentation [13]. The per class activations of the network are, therefore, easily interpretable, i.e., by a human user.

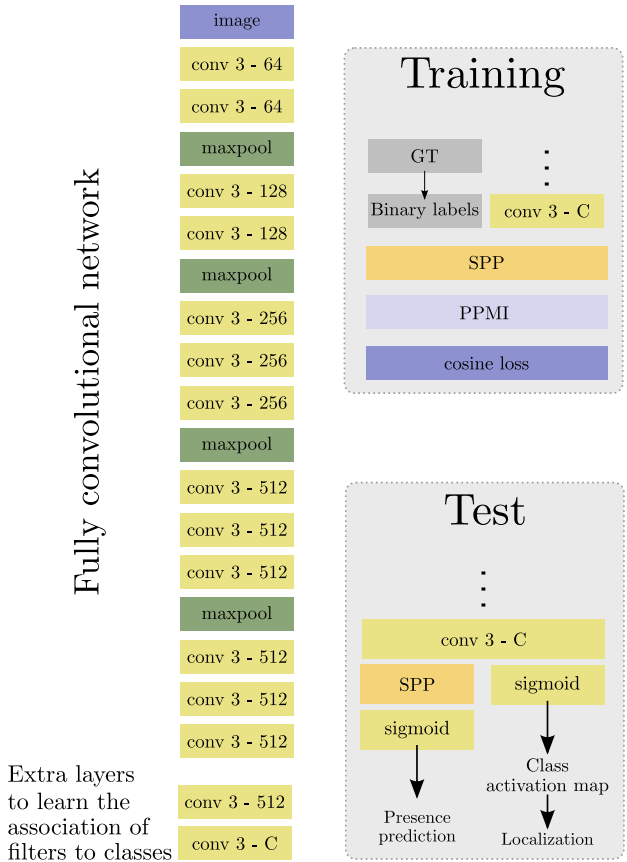

Figure 1: Overview of the proposed fully convolutional network architecture. During training both ground truth and predictions may be processed on image level or may include coarse localizations which are encoded by an Spatial Pyramid Pooling (SPP) layer. Both vectors are projected into an embedding space using the positive pointwise mutual information (PPMI) transformation which is derived from the training data. A cosine loss is computed for training. During testing, the output of the last convolutional layer can be either used for a presence prediction or a weakly supervised classification based on the class activation maps.

For a classification task this map can be processed by a pooling layer in order to indicate the presence of an object in an image. Here, it is proposed to use a SPP layer that employs average pooling. This allows to compute a presence score for a global presence but also for certain regions of an image. For training the network based on image tags or point-wise annotations, the output is projected into an embedding space using the PPMI layer. The ground truth annotations are projected to the same embedding space and then a cosine loss is computed. For evaluation, a sigmoid layer can be used in order to derive probability scores from the class activations. Weakly supervised object detection can be performed by processing the response of each pixel of the class activation map by a sigmoid function. This results in probability scores which indicate the location of objects.

### 2.2. Integrating coarse point-wise localizations

Incorporating an SPP layer in the network architecture allows for encoding additional coarse localizations for an object's presence within an image. The presence of an object can be encoded for each tile in the pyramid. Such an encoding can be combined with bounding boxes or with even simpler forms of annotations. Most interestingly, point-wise annotations allow to indicate the presence of an object in a certain region. A human annotator is asked to click on an object, therefore, indicating it's presence by a single point within the image. These point-wise annotations require a minimal manual effort [1].

As each tile indicates the presence of an object in a certain region, the SPP approach will generate different levels of granularity. Each tile that contains a point of a certain object, is labeled with this object class being present. The feature vector that is used for training is the concatenation of multiple SPP tiles. An illustration is shown in Fig. 2. Therefore, when encoding the presence with a binary vector that shall be learned by the network, multiple co-occurrences are created within this vector. In the given example, the presence of a *dog* in the upper left, as well as the upper right tile of the image at the first level of the pyramid co-occurs with the presence of a *dog* at image level. This co-occurrence will always occur for the tiles at the image level and the finer levels of detail.

### 2.3. Encoding co-occurences with pointwise mutual information

Due to the location encoding by the SPP layer, as well as the natural co-occurrences of objects in images, the binary label vectors will exhibit recurring co-occurrences. In order to take these into account, a feature space is computed that captures the likelihood that any two labels may co-occur in a given image. This feature space is then used for training the network.

Following the idea of [3], the PMI can be computed in order to measure the mutual information between labels and find correlations within the data. Here, all object occurrences within the ground truth annotations of the training

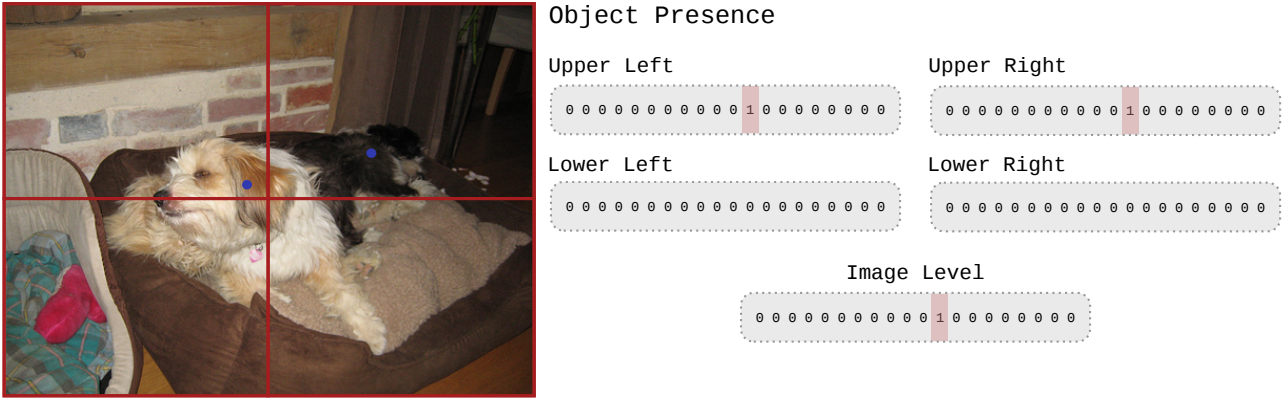

Figure 2: Illustration of the point-wise labels and the resulting co-occurrences in the resulting pyramid feature vector. Here, the two dogs are annotated with points-wise annotations (in blue) in the upper left and upper right tile (indicated in red). The resulting feature vector indicates the presence of a certain class in the respective tiles using a one. A zero indicates it's absence. The final representation is the concatenation of all tiles.

data can be used for computing

$$\text{PMI} = \left(log\frac{p(O_i, O_j)}{p(O_i)p(O_j)}\right)_{(i,j)}, \quad (1)$$

where $p(O_i, O_j)$ represents the probability of object $i$ and $j$ to occur together, $p(O_i)$ and $p(O_j)$ are the priors for object $i$ and $j$ respectively. This can either be evaluated on image level or for the complete pyramid in the SPP. In contrast to [3], the PPMI, which is defined by

$$\text{PPMI} = \max(0, PMI), \quad (2)$$

is used in the proposed approach. For object detection, the presence of objects occurring together is most important, ignoring negative correlations. This matrix can then be expressed by

$$\text{PPMI} = U \cdot \Sigma \cdot U^T \quad (3)$$

where $\Sigma$ is a diagonal matrix with the eigenvalues in the diagonal. Then, $E = U\sqrt{\Sigma}$ is considered as a transformation matrix so that the $PPMI = E \cdot E^T$. In [3] this approach was not only used for computing a transformation, but also in order to reduce dimensionality. In the presented approach, the dimensionality is preserved, which is important in the following as it allows for reconstructing the original feature vector. Note that for unobserved co-occurrences of two object classes $P(O_i, O_j)$ equals zero. Therefore, the $PMI$ matrix is not necessarily positive semidefinite so that the eigenvalues in $\Sigma$ could become negative. Without reducing the dimensionality, it is therefore imperative to use the PPMI, which yields positive semidefinite matrix, instead of PMI as otherwise $\sqrt{\Sigma}$ could become complex.

For projecting a feature vector $x$ into the embedding space, the transformation $E \cdot x$ is applied. In order to integrate this into the CNN, an additional layer is introduced that implements the embedding $E$ (see Fig. 1). The PPMI transformation is learned from the training samples before training the CNN. When training the CNN, the embedding matrix $E$ is encoded in a single fully connected layer for which the weights are not updated during the stochastic gradient descent. This layer is used in order to project the scores as well as the ground truth annotations indicating an objects presence in the image or a certain region of the SPP representation into a new embedding space. In contrast to logistic regression, where a non-continuous space (binary vectors) is used, the embedding space is continuous. Since the features in this space are in $\mathbb{R}^n$ and non-categorical, the cosine loss function is computed between the networks output $\hat{y}$ and a ground truth feature vector $y$:

$$loss(\hat{y}, y) = 1 - \frac{\hat{y}^T y}{||\hat{y}|| \cdot ||y||} \quad (4)$$

The cosine loss is chosen over the $L_2$-loss as it can be expected that distances between high-dimensional target vectors are better represented by an angle than the Euclidean distance which typically suffers from the curse of dimensionality. Moreover, the cosine loss computes to zero for an infinite amount of points for a given target while the $L_2$-loss only equals to zero for a single point (the target itself). It is reasonable to assume that this trait benefits the learning process.

When training the network with backpropagation, the PPMI layer computes a backprojection from the embedding space and reconstructs the original feature vector $x$. The error is, therefore, also evaluated with respect to the class scores.

In [3] the network is trained solely in the embedding space and thus predicts a vector in the embedding space. The presence of an attribute had therefore to be predicted based on the cosine distance between a binary vector indicating the presence of a single attribute and the PMI output. In the proposed approach, the class scores are directly obtained by a forward pass through the network.

| Annotation Detail | Loss Function | Loss Layer(s) | mAP image Level | $2 \times 2$ |
|---|---|---|---|---|
| Global | Binary logistic | – | 76.9% | 26.0% |
| Global | PPMI + Cosine | – | **80.3%** | **33.1%** |
| Global | Oquab et. al. [14] (full images) | | 76.0% | – |
| Global (*) | Oquab et. al. [14] (weakly supervised) | | **81.8%** | – |
| $2 \times 2$ | Binary logistic | Finest Level | 75.5% | 34.2% |
| $2 \times 2$ | Binary logistic | Pyramid | 77.1% | 34.4% |
| $2 \times 2$ | PPMI + Cosine | Pyramid | **82.1%** | **35.2%** |

(*) An additional multi scale analysis is carried out.

Table 1: Mean average precision for the classification in the VOC2012 dataset.

## 3. Evaluation

The proposed approach is evaluated on the VOC2012 dataset [5]. Additional coarse localizations are provided by the point-wise annotations published in [1]. These annotations were created by crowd workers, which were asked to indicate the location of an object by clicking on it.

The approach is evaluated for three tasks. First, the classification task indicating the presence of an object in an image. Second, the point-wise localization, following the setup of [14], where the highest activation in a feature map is taken in order to predict a single point indicating the location of an object. Third, the weakly supervised localization is evaluated based on the correct localization (CorLoc) accuracy [4, 10]. While, for the first two tasks, the networks are trained on the train set and evaluated on the validation set of the VOC2012 benchmark (the test set is not available for the point-wise annotations), the CorLoc metric is typically evaluated on a training set and therefore evaluated on the complete trainval split of the dataset (cf. [10]).

The training images are rescaled so that the shortest side is 512px in length. All networks are trained with a batch size of 256 for 2,000 iterations, which equals 512,000 images or 90/45 epochs on the train/trainval split. The first 600 iterations are trained with a learning rate of 0.0001 which is then increased to 0.001. Random data augmentations are applied, which include translation (up to 5%), rotation (up to 5 deg), Gaussian noise ($\sigma = 0.02$) and vertical mirroring. The networks that are trained with global, image level annotations are initialized using the ImageNet weights for the VGG16 networks. The networks which are trained with coarse point-wise localization are initialized using the weights from the image level training.

### 3.1. Classification

Table 1 reports the classification accuracy on the VOC2012 dataset as the mean average precision (mAP). The accuracy is evaluated with respect to the presence of an object in an image or in any of the tiles of a $2 \times 2$ subdivision of the image. In the latter case, each tile is evaluated independently. The prediction scores for each tile are compared to the point-wise annotations and the average over all tiles is reported.

A CNN using binary logistic loss is compared to one using the proposed combination of PPMI and a cosine loss. The results show that the performance on image level and also when using the highly correlated point-wise annotations can be improved by the PPMI embedding. It can also be seen that incorporating the coarse localizations which are derived from the point-wise annotations helps improving the image level results as well as the predictions for the more detailed $2 \times 2$ regions.

When comparing the image level results to the ones published in [14], similar results are achieved. While outperforming the *full image* setup of [14] the results are slightly below the *weakly supervised* setup which applied an additional multi-scale analysis during training. Note that our training configuration is more similar to the *full image* setup as a fixed image size is used.

### 3.2. Localization

For evaluating the localization accuracy the protocol designed in [14] is followed. The class-wise activations after the sigmoid computation are rescaled to the original image size. Using the maximum activation within a feature map, one detection point can be reported for each object class. A point prediction is considered as correct if it is within a ground truth bounding box ($\pm$ 18px) of the same class, as in [14]. Each point is then associated with its respective probability score and the mAP is computed.

The results are reported in Tab. 2. The PPMI embedding improves the results significantly compared to the binary logistic loss. Here, it can be seen that the proposed network provides a better localization result than the approach in [14]. Similar to the classification results, the additional coarse localizations allow for further improving the results at the cost of a minimal annotation effort.

| Annotation Detail | Loss Function | Loss Layer(s) | mAP localization |
|---|---|---|---|
| Global | Binary logistic | – | 69.8% |
| Global | PPMI + Cosine | – | **76.5%** |
| Global | Oquab et. al. [14] (weakly supervised) | | 74.5% |
| 2 × 2 | PPMI + Cosine | Pyramid | 78.1% |
| BBoxes | R-CNN [6]; results reported in [14] | | 74.8% |

Table 2: Results of the point-wise localization, following the setup in [14].

### 3.3. CorLoc

Last, the correct localization (CorLoc) accuracy has been evaluated. Here, the results are provided for the VOC2012 trainval set. Given an image and a target class, the CorLoc describes the percentage of images where a bounding box has been predicted that correctly localizes an object of the target class. An intersection over union (IoU) of 50% is required for a prediction to be considered as correct.

For predicting a localization, the approach of [19] is followed. Note that the network is able to predict multiple localizations for different object classes at once. Given a target class, all pixels with an activation of more than 10% of the maximum activation for the target class are chosen as foreground pixels. The bounding box around the largest connected region is chosen as the object prediction.

The results are shown in Tab. 3. The combination of PPMI and a cosine loss improves the localization by a margin compared to a binary logistic loss. Again, the coarse localizations are able to produce more precise results. Note that recently an approach has been proposed that achieves a CorLoc of $54.8\%$ by incorporating selective search data and explicitly optimizing for bounding box predictions [10]. In contrast, the proposed approach is trained in an end-to-end fashion without additional selective search data.

### 3.4. Qualitative results

Exemplary results are shown in Fig. 3. The examples are taken from the CNN that has been trained for the CorLoc on the VOC2012 trainval set. The network has been trained using the proposed combination of PPMI and a cosine loss. The annotations are coarse localizations derived from point-wise annotations for 2 × 2 tiles. The left column shows the input image with the bounding boxes of the target class shown in green. The middle column shows the predicted object region after thresholding the class activation map. The right side shows the class activation map as derived from the network. The class activation map is the output of a single feature map where each pixel's intensity has been processed by a sigmoid function.

It can be observed that the desired objects are nicely localized in the feature maps. Even in the error case, the activations are reasonable as multiple screens are placed close to each other, making it difficult to distinguish them in a weakly supervised fashion.

## 4. Conclusion

In this work a novel approach for weakly supervised object detection with CNNs has been proposed. The network incorporates the positive pointwise mutual information (PPMI) and a cosine loss function for learning. It is shown that this approach eases the learning process, improving the results compared to a binary logistic loss based on categorical feature vectors.

A fully convolutional network architecture has been used for the weakly supervised detection. A single feature map is learned for each class, creating an intuitive representation for the presence of an object class in an image. Furthermore, an SPP layer is incorporated in the network instead of a global pooling operation. This allows for incorporating coarse localizations, i.e., in the form of point-wise annotations. These annotations require a minimal manual effort, but provide additional information that can be leveraged for weakly supervised localization.

The evaluation on the VOC2012 dataset shows that the combination of PPMI and a cosine loss improves the results for classification, point localization as well as the CorLoc. Furthermore, the additional point-wise annotations helps in steering the learning process and further improve the results for all three tasks at a minimal annotation cost.

## 5. Acknowledgment

This work has been supported by **** an anonymous institution *** .

| Annotation Detail | Loss Function | Loss Layer(s) | Initialization | CorLoc |
|---|---|---|---|---|
| Global | Binary logistic | – | ImgNet | 26.2% |
| Global | PPMI + Cosine | – | ImgNet | **39.2%** |
| Global (*) | Kolesnikov et. al. [11] | | | **54.8%** |
| 2 × 2 | PPMI + Cosine | Pyramid | SPL1 | **43.4%** |

(*) Requires additional selective search data

Table 3: CorLoc on the VOC2012 trainval set with an IoU of 50%.

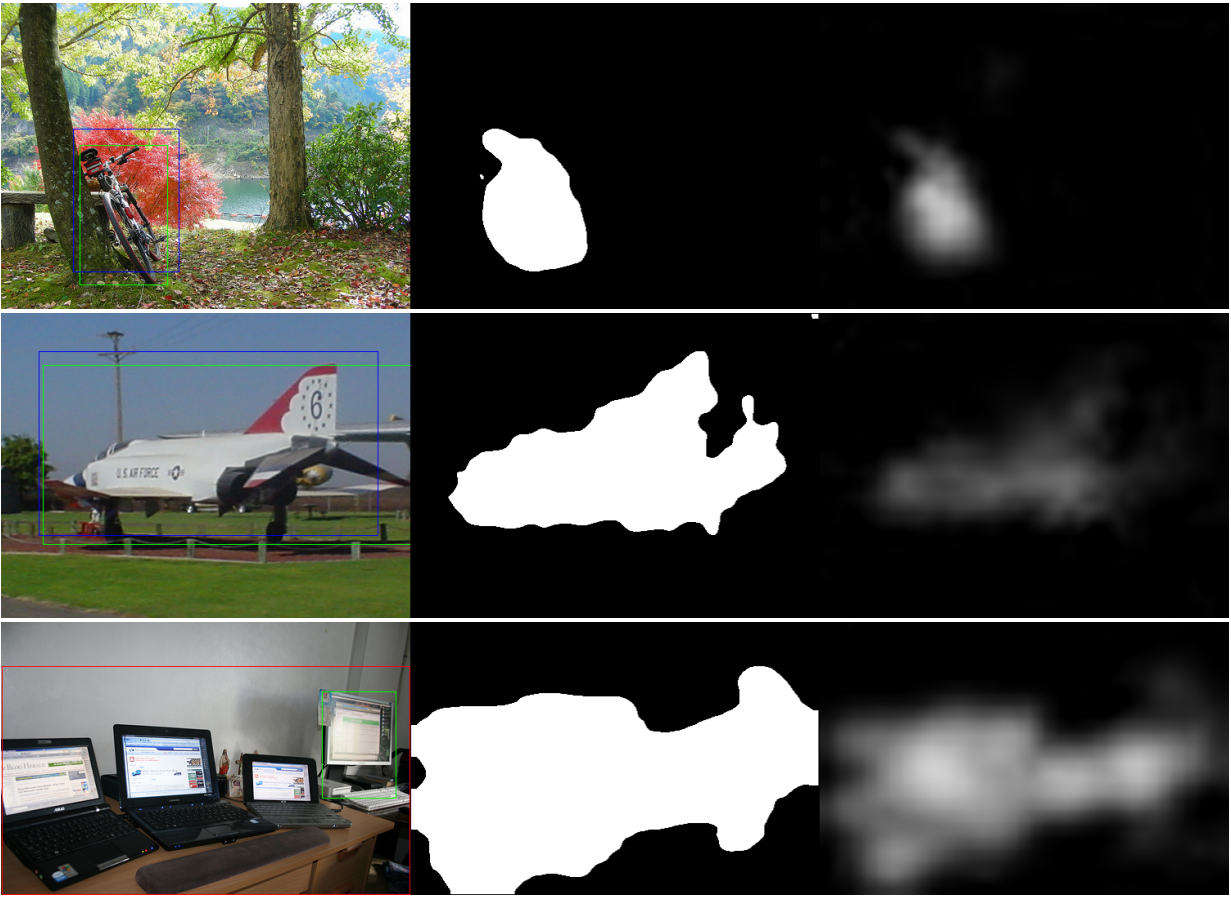

Figure 3: Qualitative results for the best performing network using 2 × 2 tiles for coarse localizations, derived from point-wise annotations. (left) Original images with annotated bounding boxes in green, predicted bounding boxes in blue and red for correct and incorrect predictions respectively. (middle) thresholded activations as used for bounding box computation. (right) class activation map as computed by the CNN.